\documentclass{sigchi}

\toappear{\scriptsize Permission to make digital or hard copies of all or part of this work for personal or classroom use is granted without fee provided that copies are not made or distributed for profit or commercial advantage and that copies bear this notice and the full citation on the first page. Copyrights for components of this work owned by others than ACM must be honored. Abstracting with credit is permitted. To copy otherwise, or republish, to post on servers or to redistribute to lists, requires prior specific permission and/or a fee. Request permissions from permissions@acm.org. \\
{\emph{CHI '20, April 25--30, 2020, Honolulu, HI, USA.} } \\
\copyright~2020 Association for Computing Machinery. \\
ACM ISBN 978-1-4503-6708-0/20/04\ ...\$15.00. \\
http://dx.doi.org/10.1145/3313831.3376544 }

\usepackage{balance}       
\usepackage{graphics}      
\usepackage[T1]{fontenc}   
\usepackage{txfonts}
\usepackage{mathptmx}
\usepackage[pdflang={en-US},pdftex]{hyperref}
\usepackage{xcolor}
\usepackage{booktabs}
\usepackage{textcomp}

\usepackage{amsmath}
\usepackage[gen]{eurosym}
\usepackage{cuted}
\usepackage{capt-of}

\usepackage{microtype}        
\usepackage{ccicons}          

\usepackage{todonotes}

\def\plaintitle{GAZED-- Gaze-guided Cinematic Editing of Wide-Angle Monocular Video Recordings}

\def\emptyauthor{}
\def\plainkeywords{Authors' choice; of terms; separated; by
  semicolons; include commas, within terms only; this section is required.}

\makeatletter
\def\url@leostyle{%
  \@ifundefined{selectfont}{
    \def\UrlFont{\sf}
  }{
    \def\UrlFont{\small\bf\ttfamily}
  }}
\makeatother
\def\pprw{8.5in}
\def\pprh{11in}

\definecolor{linkColor}{RGB}{6,125,233}
\hypersetup{%
  pdftitle={\plaintitle},
  pdfauthor={\emptyauthor},
  pdfkeywords={\plainkeywords},
  pdfdisplaydoctitle=true, 
  bookmarksnumbered,
  pdfstartview={FitH},
  colorlinks,
  citecolor=black,
  filecolor=black,
  linkcolor=black,
  urlcolor=linkColor,
  breaklinks=true,
  hypertexnames=false
}
\begin{document}

\def\x{{\mathbf x}}
\def\L{{\cal L}}
\def\eg{\textit{e.g.}}
\def\ie{\textit{i.e.}}
\def\Eg{\textit{E.g.}}
\def\etal{\textit{et al.}}
\def\etc{\textit{etc}}
\newcommand{\blue}[1]{\textcolor{black}{#1}}
\newcommand*{\Perm}[2]{{}^{#1}\!P_{#2}}%
\newcommand*{\Comb}[2]{{}^{#1}C_{#2}}%

\title{\plaintitle}

\numberofauthors{4}
\author{%
  \alignauthor{K L Bhanu Moorthy\\
    \affaddr{CVIT, IIIT Hyderabad}\\
    \affaddr{Hyderabad, India}\\
   \email{k.l.bhanu@research.iiit.ac.in}}\\ 
  \alignauthor{ Moneish Kumar \thanks{Work done while at IIIT Hyderabad}\\
    \affaddr{Samsung R\&D Institute}\\
    \affaddr{Bangalore, India}\\
    \email{moneish04@gmail.com}}\\ 
  \alignauthor{Ramanathan Subramanian\\
    \affaddr{IIT Ropar, India}\\
   \email{s.raamanathan@iitrpr.ac.in}}\\ 
      \alignauthor{Vineet Gandhi\\
    \affaddr{CVIT, IIIT Hyderabad}\\
    \affaddr{Hyderabad, India}\\
   \email{vgandhi@iiit.ac.in}}\\ 
} 
  
\teaser{\centering
\includegraphics[width=17cm]{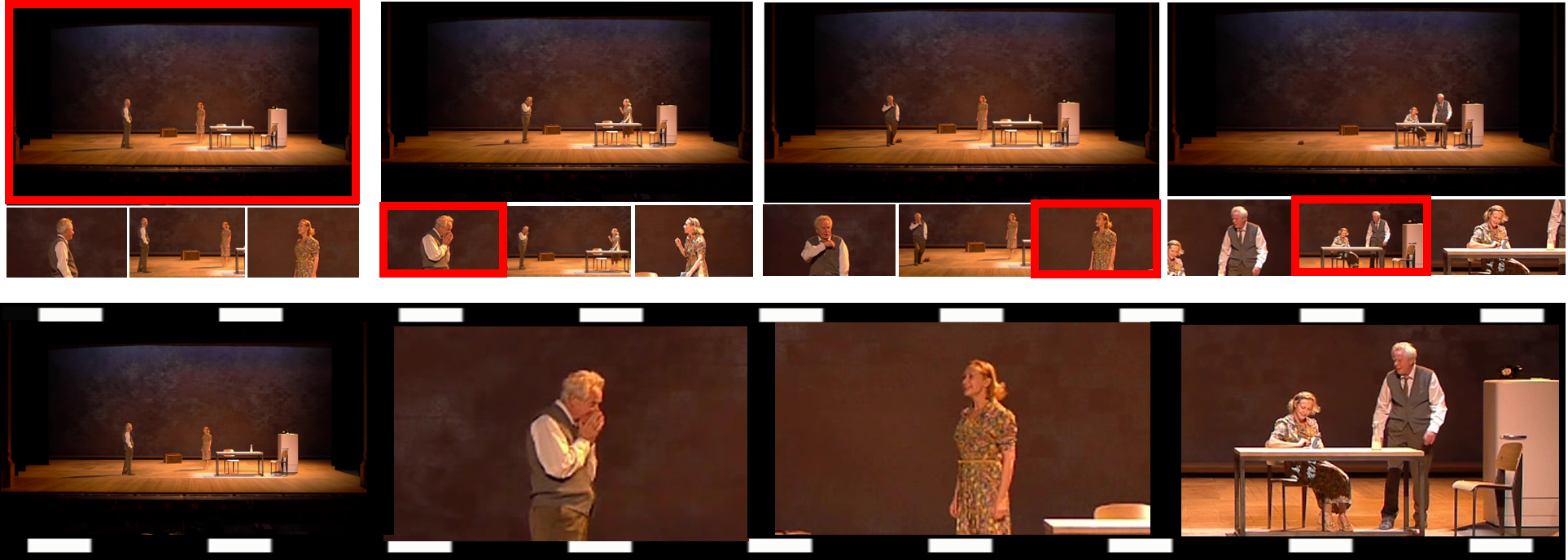}
\captionof{figure}{We present GAZED: gaze-guided and cinematic editing of monocular and static stage performance recordings. (Top row) GAZED takes as input frames from the original video (\textit{master shots}), and generates multiple \textit{rushes} by simulating several virtual Pan/Tilt/Zoom cameras. Generated rushes are shown below each frame. Eye gaze data are then utilized to \textit{select} the \textit{best} rush (shown in red) for each master shot. The \textit{edited video} (a) vividly presents scene emotions and actions, and (b) adheres to cinematic principles to present aesthetic content. {Edited video} frames are shown in the bottom row. Best viewed in color.
\label{fig:teaser}}
} 

\maketitle

\begin{abstract}
We present \textbf{GAZED}-- {eye GAZe-guided} EDiting for videos captured by a solitary, static, wide-angle and high-resolution camera. Eye-gaze has been effectively employed in computational applications as a cue to capture \textit{interesting} scene content; we employ gaze as a proxy to \textit{select shots} for inclusion in the \textit{edited video}. Given the original video, scene content and user eye-gaze tracks are combined to generate an edited video comprising \textbf{\textit{cinematically valid}} \textit{actor shots} and \textit{shot transitions} to generate an aesthetic and vivid representation of the original narrative. We model cinematic video editing as an \textit{energy minimization problem} over \textit{shot selection}, whose constraints capture cinematographic editing conventions. Gazed scene locations primarily determine the shots constituting the edited video. Effectiveness of GAZED against multiple competing methods is demonstrated via a psychophysical study involving 12 users and twelve performance videos.
\end{abstract}



\section{Introduction}
Professional video recordings of {stage performances} are typically created by employing skilled camera operators, who record the performance from multiple viewpoints. These multi-camera feeds, termed \textit{rushes}, are then \textit{edited} together to portray an eloquent story intended to maximize viewer engagement. Generating professional edits of stage performances is both difficult and challenging. Firstly, maneuvering cameras during a live performance is difficult even for experts as there is no option of retake upon error, and camera viewpoints are limited as the use of large supporting equipment (trolley, crane~\etc.) is infeasible. Secondly, manual video editing is an extremely slow and tedious process and leverages the experience of skilled editors. Overall, the need for (i) a professional camera crew, (ii) multiple cameras and supporting equipment, and (iii) expert editors escalates the process complexity and costs. 


Consequently, most production houses employ a large field-of-view static camera, placed far enough to capture the entire stage. This approach is widespread as it is simple to implement, and also captures the entire scene. Such static visualizations are apt for archival purposes; however, they are often unsuccessful at captivating attention when presented to the target audience. While conveying the overall context, the distant camera feed fails to bring out vivid scene details like close-up faces, character emotions and actions, and ensuing interactions which are critical for cinematic storytelling (Fig.~\ref{fig:teaser}, top row). Renowned film editor Walter Murch states in his book~\cite{murch2001blink} that a primary objective of film editing is to register the expression in the actor's eyes. If this is not possible, one should attempt to capture the best possible close-up of the actor, even if the original wide-shot may convey the scene adequately.

GAZED denotes an end-to-end pipeline to generate an \textit{aesthetically edited video} from a single static, wide-angle stage recording. This is inspired by prior work~\cite{gandhi2014multi}, which describes how a plural camera crew can be replaced by a \textit{single} high-resolution static camera, and \textit{multiple virtual camera shots} or \textit{rushes} generated by simulating several virtual pan/tilt/zoom cameras to focus on actors and actions within the original recording. In this work, we demonstrate that the multiple rushes can be automatically edited by leveraging user \textit{eye gaze} information, by modeling (virtual) \textit{shot selection} as a discrete optimization problem. Eye-gaze represents an inherent guiding factor for video editing, as eyes are sensitive to \textit{interesting} scene events~\cite{Ramanathan09,Subramanian2014} that need to be vividly presented in the edited video.

Bottom insets in the top row of Fig.~\ref{fig:teaser} show the multiple rushes (which are either manually shot or virtually generated)  to choose from at any given time; the objective critical for video editing and the key contribution of our work is to \textit{decide which shot (or rush) needs to be selected to describe each frame of the edited video}. The shot selection problem is modeled as an optimization, which incorporates gaze information along with other cost terms that model \textit{cinematic editing principles}. Gazed scene locations are utilized to define \textit{gaze potentials}, which measure the {importance} of the different shots to choose from. Gaze potentials are then combined with other terms that model cinematic principles like \textit{avoiding jump cuts} (which produce jarring shot transitions), \textit{rhythm} (pace of shot transitioning), \textit{avoiding transient shots}~\etc. The optimization is solved using \textbf{\textit{dynamic programming}}~(\url{https://en.wikipedia.org/wiki/Dynamic_programming}). 

To validate GAZED, we compare multiple edited versions of 12 performance recordings via a psychophysical study involving 12 users. Our editing strategy outperforms multiple competing baselines such as \textit{random} editing, \textit{wide-shot} framing, \textit{speaker detection}-based editing and \textit{greedy gaze}-based editing. Our contributions include:

(1) \textbf{Gaze potential for shot selection:} We model user eye-gaze information via gaze potentials that quantify the \textit{importance} of the different shots (rushes) generated from the original recording. Our algorithm examines locations as well as the extent of gaze clusters in a bottom-up fashion to compute unary and higher-order gaze potentials. Human gaze is known to be more sensitive to high-level scene semantics such as emotions~\cite{Ramanathan09,Subramanian2014}, as compared to bottom-up computational saliency methods.

(2) \textbf{Novel video editing methodology:} We perform shot selection by minimizing an objective function modeled via gaze potentials plus constraints conveying cinematic principles. GAZED edits a 1-minute video with four performers on stage in 5 seconds on a \blue{PC with 7th generation Intel 2.7~GHz i5 processor and 8GB RAM}. In contrast, manual editing is both time and effort-intensive. 

(3) \textbf{An end-to-end cinematic editing pipeline:} to generate professional videos from a static camera recording. \blue{GAZED enables novice users to create professionally edited videos using their eye gaze data and an inexpensive desktop eye-tracker. The \textit{Tobii Eye-X} eye-tracker costing $<\EUR{200}$ was used to compile eye gaze data in this work.} We employ the algorithm of Gandhi \etal~\cite{gandhi2014multi} to generate multiple rushes, followed by dynamic programming to obtain the edited video. 

(4) \textbf{A thorough user study:} to validate the superiority of GAZED against multiple editing baselines. Results confirm that GAZED outputs are preferred by users with respect to several attributes characterizing editing quality. 

\section{Related Work}

\subsection{Editing in virtual 3D environments}
The video editing problem has been extensively studied in virtual 3D environments, where cinematographic conventions are computationally modeled to best convey the animated content. Earlier works~\cite{christianson1996declarative, he1996virtual} focused on an idiom based approach. Film idioms~\cite{arijon1976grammar} are a collection of stereotypical formulas for capturing specific scenes as sequences of shots. Difficulty to formulate every specific scenario significantly limits idiom-based approaches. Moreover, in constrained settings like live theatre, it may not always be feasible to acquire all the ingredient shots of the formula. 

Another series of papers pose video editing as a discrete optimization problem, solved using dynamic programming~\cite{elson2007lightweight, lino2011computational, galvane2015continuity,  merabti2016virtual}. The key idea is to define the importance of each shot based on the narrative, and solve for the best set of shots that maximize viewer engagement. The work of Elson \etal~\cite{elson2007lightweight} couples the twin problems of camera placement and camera selection; however, details are insufficiently described to be reproduced. Meratbi \etal~\cite{merabti2016virtual} employs a Hidden Markov Model for the editing process, where shot transition probabilities are learned from existing films. They only limit to dialogue scenes, which require manual annotation from actual movies. The work of Galvane \etal~\cite{galvane2015continuity} decouples the camera positioning and selection problems, and is arguably the most comprehensive effort in video editing, precisely addressing several important issues like exact placement of cuts, rhythm and continuity editing rules. While our work is inspired by these efforts, stage performances are significantly constrained as neither is there freedom to freely place  cameras, nor does one have access to the precise scene geometry, character localization, events and action information which is available in 3D environments on which the above works are applicable. 

\subsection{Editing 2D videos}
Automated video editing has also been studied in other specialized scenarios. The Virtual Videography system~\cite{heck2007virtual} simulates shots captured by virtual pan-tilt-zoom cameras from lecture videos, and the best shot is selected via branch-and-bound optimization. The MSLRCS~\cite{zhang2008automated} and AutoAuditorium~\cite{bianchi2004automatic} systems use a small set of fixed cameras including input of presentation slides. Editing is done via a rule-based approach limited to a constrained environment of usually a single presenter in front of a chalkboard/slide screen. 

Ranjan \etal~\cite{ranjan2008improving} propose a system for editing group meetings based on speaker detection cues, posture changes, and head orientation. They model editing in the form of simple rules like \textit{cut to the close up of the speaker} when a speaker change is detected and \textit{cut to an overview shot} when multiple people are speaking. In our experiments, we compare GAZED with a similar baseline which focuses on the speaker(s) over time. Doubek \etal~\cite{doubek2004cinematographic} study the problem of camera selection in surveillance and telepresence settings. 

The camera selection problem has been extensively studied in sports events~\cite{wang2008automatic,chen2010personalized,ChenWHCSG13,chen2018camera}. Earlier works~\cite{wang2008automatic,chen2010personalized} employ Hidden Markov Models to select from within a panoramic view, or from multiple views. Other works~\cite{ChenWHCSG13,chen2018camera} take a data-driven approach and train regressors to learn the importance of each camera view at a given time. A recent interesting work by Leake \etal~\cite{leake2017computational} proposes an idiom based approach for editing dialogue-driven scenes. Inputs to their system include multiple camera views with the film script, and the output is the most informative set of shots for each line of the dialogue. In contrast to the approaches which work on multi-camera feeds plus scene-related meta-data, GAZED only requires a static wide-angle camera recording of a stage performance, and eye gaze data from one or more viewers (with a high-end eye-tracker, eye gaze recording of the editor/director reviewing the event would be sufficient). Our method does not rely on script alignment, multiple carefully placed cameras, speaker information \etc. Furthermore, our approach is applicable to a wide variety of scenarios, and we show results on \textit{theatre}, \textit{dance} and \textit{music} performances as part of the supplementary video.

\subsection{Gaze driven editing}
\begin{figure}[t]
    \centering
   \includegraphics[width=\linewidth]{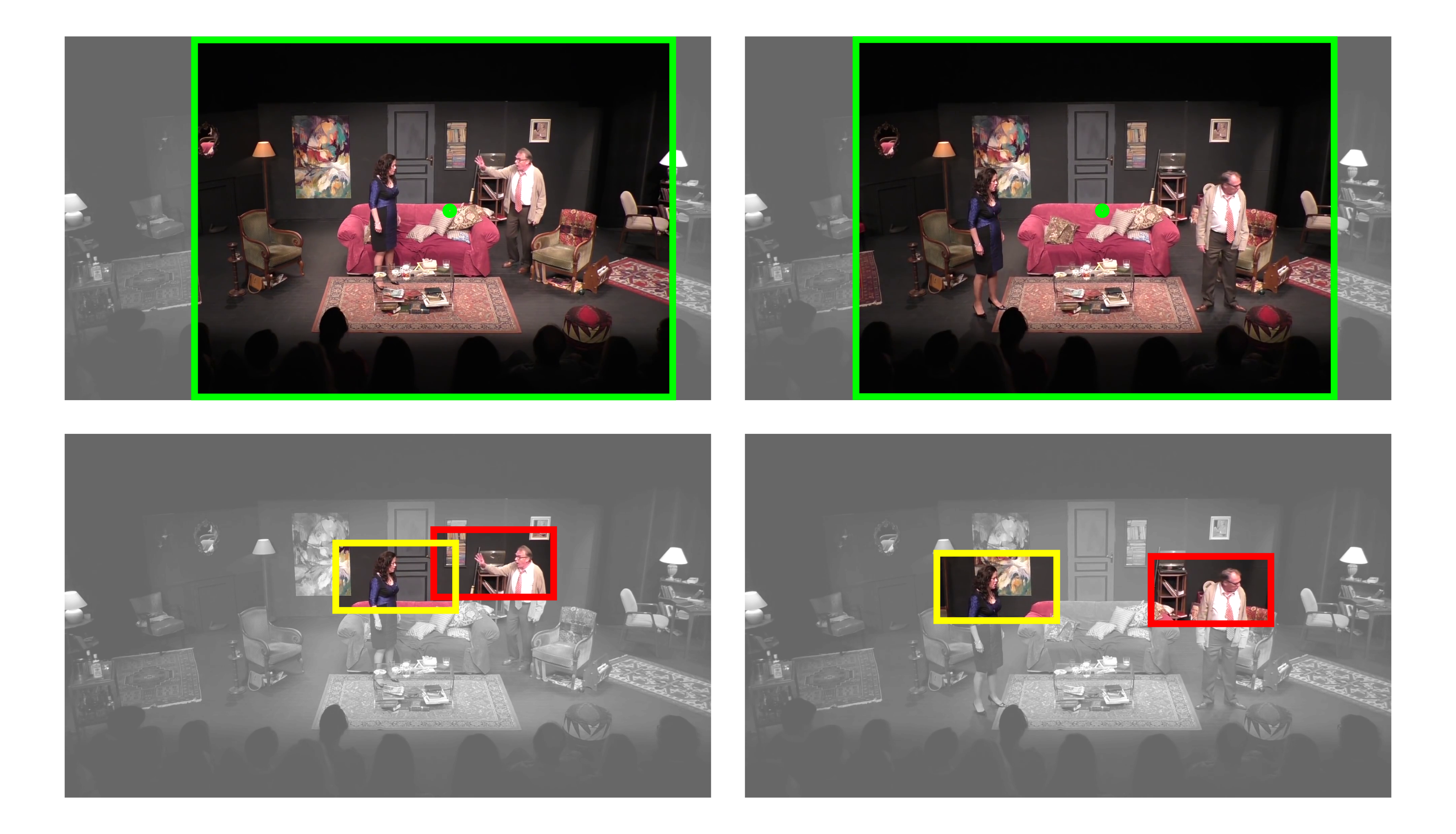} 
    \caption{Video retargeting vs. editing: \textit{Video retargeting} (top row) involves estimating the location of a cropping window (defined by the $x$-coordinate denoted by a green dot) that preserves focal scene content for rendering on a device with different aspect ratio. \textit{Video editing} (bottom row) involves compositing several manually captured or virtually generated shots (two of the simulated shots are shown) to capture focal characters in vivid detail, and selecting at each instant the shot that most engagingly conveys the storyline. Selected shots are rendered at the same aspect ratio as the original video. }\vspace{-.2cm}
    \label{retargeting-vs-editing}
\end{figure}
Prior works have also employed eye-gaze for video editing. The gaze is estimated either via head pose or through specialized eye-tracking devices. Takemai \etal~\cite{takemae2004impact} propose a video editing system based on the participant's gaze direction in indoor conversations. They show results on debate scenarios comprising 3--4 participants, where participant gaze is manually labelled for the entire video. They apply a simple editing rule, \ie, to cut to the close-up of the person that most peers are gazing at. They demonstrate that gaze is able to better convey the flow of conversation compared to a speaker detection based approach. Daigo \etal~\cite{daigo2004automatic} use audience gaze direction to estimate the areas of interest in a basketball match. Park \etal~\cite{park20123d} and Arev \etal~\cite{arev2014automatic} identify salient scene regions from the convergence of the field-of-view of multiple social cameras viewing the scene.  

Two recent works~\cite{jain2015gaze, rachavarapu2018watch} employ eye gaze data for the problem of \textit{video retargeting}, which is meant to \textit{adapt} an \textit{edited video} designed for a specific display device to a different one (\eg, theatre screen to a mobile device). This is often achieved by virtually moving a cropping window within the original video, which preserves the salient scene content (Fig.~\ref{retargeting-vs-editing}). These methods primarily solve for an $x$ value at each time point, and allow nominal zoom based on gaze variance. Since these methods do not solve for $y$, they cannot be used to obtain well-composed framings (\eg, medium shot, close-up) of the scene objects and actions. Substantial zoom in ~\cite{jain2015gaze, rachavarapu2018watch} ends up generating odd compositions (frame covering head but not face; actors cropped in an unaesthetic way, \etc). In contrast, the proposed work emulates the multi-camera video production-plus-editing pipeline, and focal scene actors and events are captured via efficient shot selection. GAZED performs shot selection among several \textit{rushes} (which are either part of the original capture or virtually generated), where each rush is carefully composed based on cinematic rules and is parameterized by the $x$, $y$ values denoting the cropping window location, along with the zoom level. In summary, the GAZED system is designed to accomplish the (edited) video creation process (not just adaptation), and can analogously work with a multi-camera setting capturing a wide scene.




\section{GAZED Overview}



GAZED represents an end-to-end system to automate the entire video production process for staged performances. The system takes as input a static, high-definition recording of the scene and eye-tracking data of one or more users who have reviewed the video recording (typically an expert such as the program director/editor). It then outputs an edited video that adheres to common cinematic principles and is aesthetically pleasing to watch. Like the traditional video production process, we also split the task into two parts (i) The \textit{generation} of cinematically valid shots capturing context as well as actions and facial expressions of performers. (ii) \textit{Selection} of the most appropriate shot at each time for engaging story-telling.




\subsection{Shot Generation} \label{shot_simulation}

GAZED takes as input a wide-angle recording captured with a static camera, covering the entire scene. We term each frame in this input wide-angle video as the `master shot' in the rest of the paper. Our work assumes that tracks (bounding box coordinates) of all the performers/actors are available for all master shots. Our approach is agnostic to the object detection/tracking algorithm; however, in this work, we employ OpenPose~\cite{cao2018openpose} as a person detector. Hungarian matching was then employed with the pairwise appearance and proximity costs for generating identity-preserving actor tracks. Tracking errors (if any) were manually corrected. A virtual camera simulation approach~\cite{gandhi_ronfard_gleicher_2014} was then used to automatically generate multiple zoomed-in shots from the original wide-angle recording. The method simulates multiple virtual pan-tilt-zoom (PTZ) cameras by moving multiple cropping windows (each following a particular actor or a group of actors) inside the master shot.  The shot generation problem is cast as a convex optimization, which takes into consideration \textit{composition} (framing/positioning of actors in the frame), \textit{panning} and \textit{cutting} aspects of cinematography.The optimization translates noisy shot estimates into well-composed cinematic shots similar to those generated by professional cameramen.




\begin{figure}[t]
    \centering
   \includegraphics[width=\linewidth]{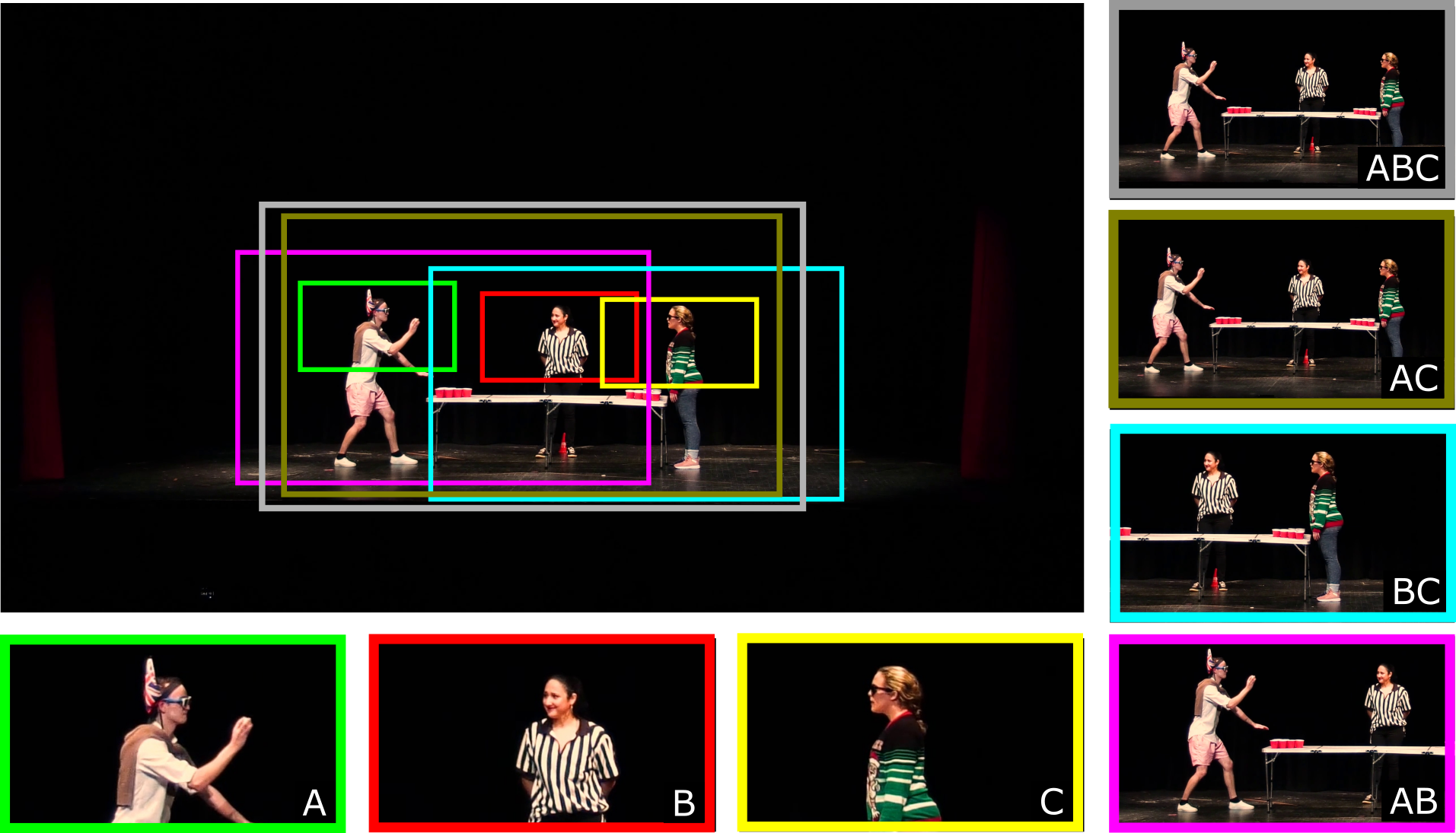} 
    \caption{Virtual shot generation: Rushes simulated from original video frame are denoted via colored rectangles. Anti-clockwise from top left: \textit{Master shot} with actors A, B and C. Generated 1-shots for each actor. Three 2-shots (AB, BC, AC) and a 3-shot (ABC comprising all three actors).}
    \label{fig:shot_generation}
    \vspace{-.3cm}
\end{figure}

Given an input video, we first generate the set of all possible shots for every combination of performers in the video. For a video with $n$ performers, there are $2^n-1$ combinations possible. Figure~\ref{fig:shot_generation} illustrates an exemplary video frame, and the set of constituent shots, which include three 1-shots (individual actor shots), three 2-shots (shots containing two actors) and a 3-shot (shot containing all actors). Similarly, for an $N$ actor sequence we generate $\Comb{n}{1}$  1-shots; $\Comb{n}{2}$ 2-shots; $\Comb{n}{3}$  3-shots and so on. The master shot along with the generated shots are referred to as  \textit{rushes}, on which editing or shot selection has to be performed. Each of the generated shots $S=\{s^i\}_{i=1}^{2^n-1}$ is parameterized by its center $c_t^i = \{x_t^i, y_t^i\}$ and width $w_t^i$ at each video time instant $t = [1 ... T]$.

\begin{figure*}[t!]
    \centering
   \includegraphics[width=0.92\textwidth]{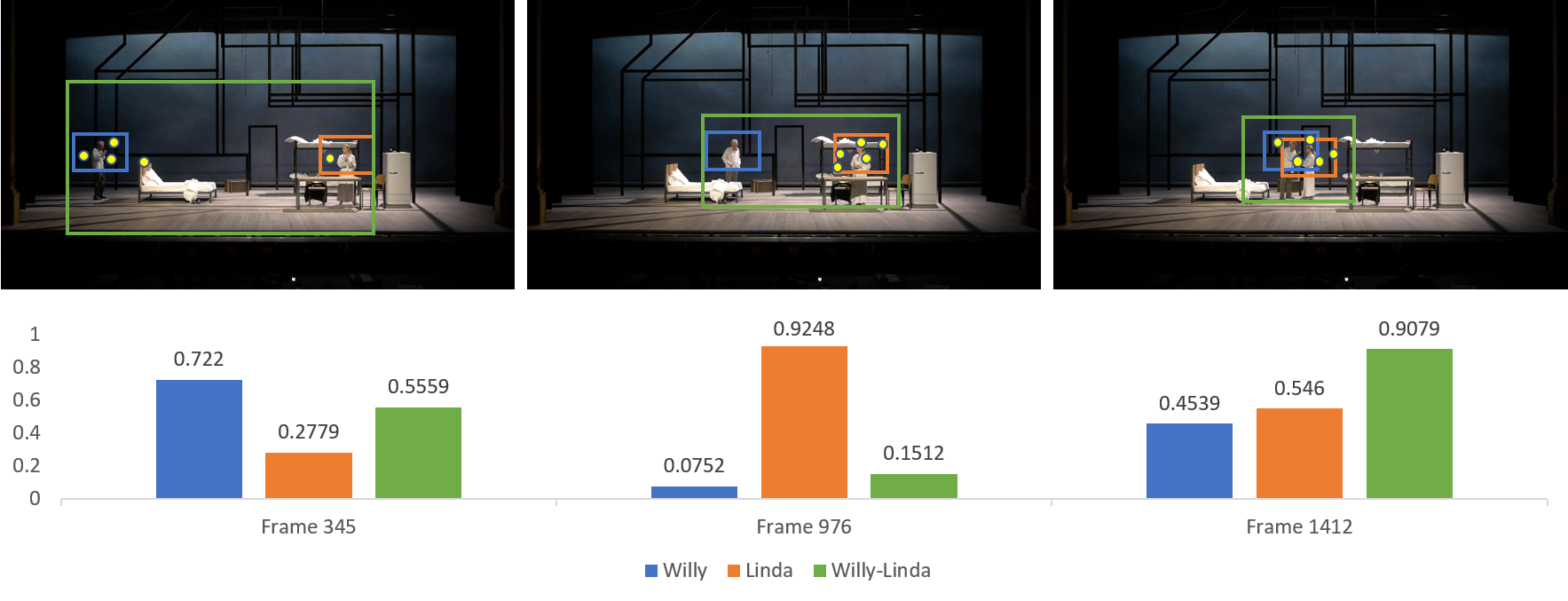} 
    \caption{Computation of gaze potentials: The first row illustrates three different frames from the \textit{master shot} and corresponding user gaze points (yellow dots). Generated rushes are denoted via colored rectangles (two \textit{one}-shots and one \textit{two}-shot in each frame). Second row presents gaze potential for each rush in the frame. Note that the gaze potential is higher for 1-shots when gaze points are concentrated (middle frame), and higher for the 2-shot when attention is dispersed (left and right frames).}
    \label{fig:gaze_potential2}\vspace{-.3cm}
\end{figure*}

We use a Medium Shot (MS) or a Medium Close-Up (MCU) for generating single actor shots (1-shots). A medium shot captures a performer from head to waist, and a medium close-up goes from head to mid-chest. Smaller framings such as the MCU and MS present an actor in vivid detail and enable viewers to better focus on the character's actions and expressions. We use a Full Shot (FS) for multiple actor sequences (2-shots, 3-shots \etc). A full shot for multiple performers is defined such that it captures each performer from head to toe. Larger framings such as the FS allow for a good deal of context around an actor to be included in the shot; these shots help establish the location where a subject is present and his/her interaction with the surroundings.

\subsection{Shot Selection}\label{shot_selection}

Given the multiple rushes, the next step involves the selection of the shot that most vividly narrates the storyline at each time instant. The GAZED algorithm poses shot selection as a discrete optimization problem, which examines the \textit{importance} of each of the multiple shots generated for every video frame, while adhering to cinematic principles like avoiding cuts between overlapping shots (termed \textit{jump cuts}), avoiding \textit{rapid shot transitions}, maintaining a \textit{cutting rhythm}, \etc. The importance of each shot at a given time is computed from eye gaze data collected using an eye-tracking device. Cinematic principles are modeled in the form of penalty terms. The final solution is obtained via a search for the optimal path through an editing graph~\cite{galvane2015continuity}. For a scene with $n$ actors, the editing graph consists of $2^n-1$ nodes at each frame (time step) $t$, where each node represents a rush and edges across time steps represent a transition from one shot to another (denoting a cut) or to itself (no cut). 

Formally, given a sequence of frames $t=[1..T]$, the set of generated shots (rushes) $S_t=\{s_t^i\}_{i=1}^{2^n-1}$ and the raw gaze data $g_t^k$ corresponding to user $k$ at frame $t$, our algorithm selects a sequence of shots $\epsilon = \{r_t\}, r_t\in S_t$ for each frame $t$, minimising the following objective function:

\vspace{-.2cm}
\begin{equation} \label{objective_function}
E(\epsilon) = \sum_{t=1}^T -ln(G(r_t)) + \sum_{t=2}^T E_e(r_{t-1},r_t)
\vspace{-.1cm}
\end{equation}

\noindent where $G(r_t)$ is a unary cost that represents the \textit{gaze potential} (modeling importance) for each shot, and $E_e(r_{t-1},r_t)$ denotes cost for switching from one shot to another.

\subsubsection{Gaze potential}

A well-edited video should engagingly convey the original narrative in a scene. Hence, it is important that the selection algorithm prefers shots which best showcase the \textit{focal} scene events at any given time. To incorporate this idea into GAZED, we quantitatively measure the importance of each rush at every time instant. Previous works (\cite{leake2017computational},\cite{galvane2015continuity}) estimate the actions/emotions in a given shot by either relying on additional meta-data or bottom-up computational features, which do not account for high-level scene semantics which humans are sensitive to. Jain~\etal~ \cite{jain2015gaze} and Rachavarapu~\etal~ \cite{rachavarapu2018watch} have shown that the gaze data recorded from users enables  effective localization of focal scene events. We extend this idea and propose a novel method to calculate the gaze potential of each shot using gaze data from users. 


The \textit{point-of-gaze} $g_t^k$ of user $k$ at each frame $t$ is defined by the $(x,y)$ coordinates where the user is looking. To determine the gaze potential for each shot, we adopt a bottom-up approach, in which we first calculate the gaze potential of lower-order 1-shots, which capture individual actors. Gaze potential for shots with multiple actors (higher-order shots) are then computed from the gaze potentials of constituent lower-order shots.     




We first compute the distance-to-center for all gaze points in a given frame $t$, considering 1-shots. The distance measure for a shot $s_i$ at time $t$ with frame center $c^i_t$ is defined as $d^i_t = \sum_{k} ||c_t^i - g_t^k||_2$. If gaze points are clustered around a particular shot (actor), $d^i_t$ will be small, while $d^i_t$ will be larger if gaze is dispersed away from the shot center. We then use the distance measure $d^i_t$ to compute gaze potentials for 1-shots as $G(s_t^i) = \frac{1}{d_t^i}/\sum_{i}\frac{1}{d_t^i}$.
The denominator is summed only over the 1-shots. The function returns a higher potential (degree of importance) for shots with focused gaze clusters, and lower potential for shots with dispersed gaze points. Fig.~\ref{fig:gaze_potential2} presents the computation of gaze potentials for an exemplar shot. We can observe that the 1-shot attracting greater user attention has higher gaze potential. 





        
        
            
            
                
        

Now, consider two 1-shots $s^a_t$ and $s^b_t$ at frame $t$, and their gaze potentials $G(s^a_t)$ and $G(s^b_t)$. Gaze potential $G(s^{ab}_t)$ of the \textit{combined} {\textit{2-shot}} for actors $a$ and $b$ is given by:
\begin{equation} \label{gaze_potential2}
G(s_{t}^{ab}) = G(s_{t}^{a}) + G(s_{t}^{b}) - | G(s_{t}^{a}) -G(s_{t}^{b})|
\end{equation}

\begin{figure}[t]
    \centering
   \includegraphics[width=0.82\linewidth]{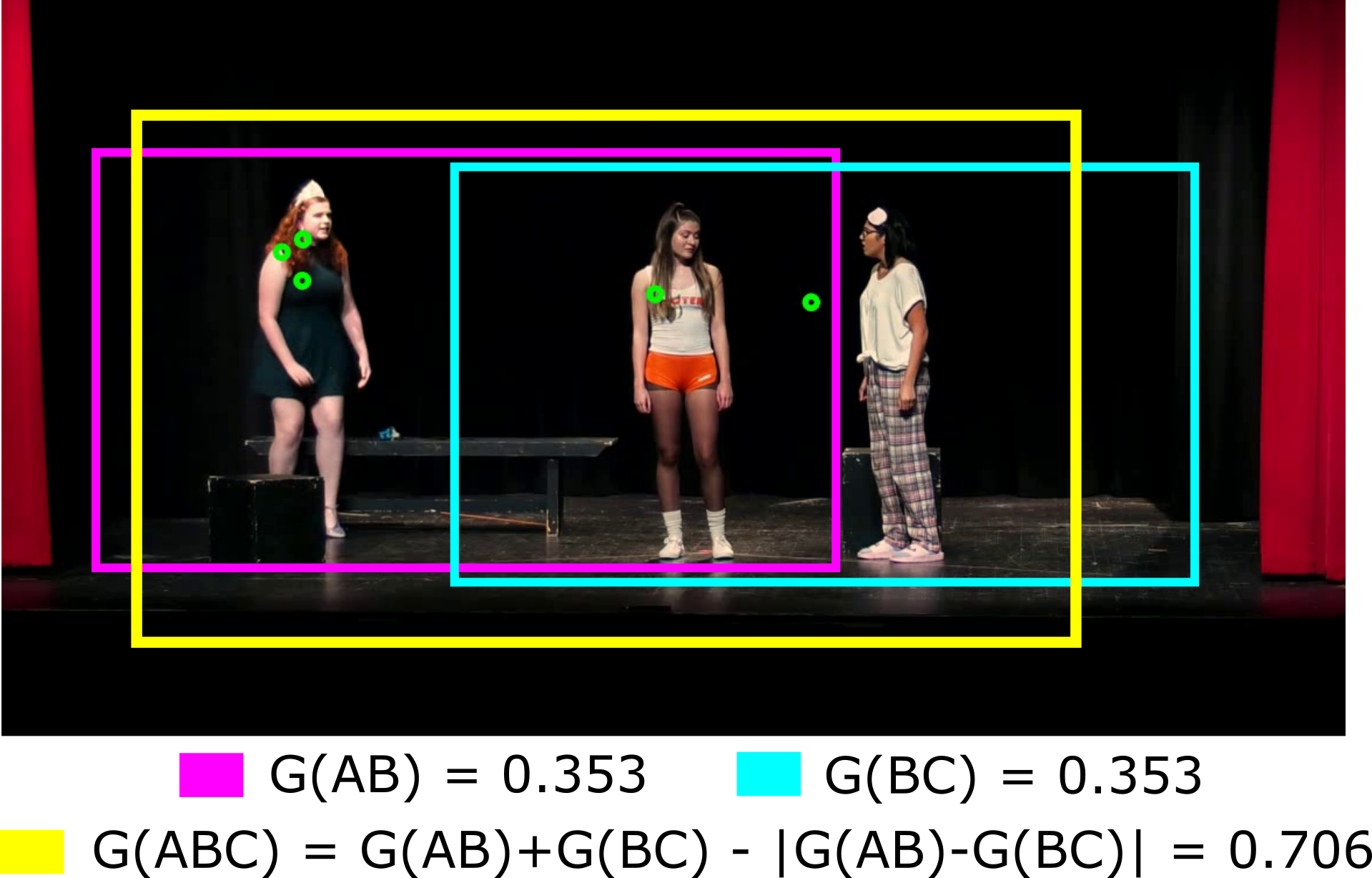} 
    \caption{Actors are ordered from left to right as A,B,C. Gaze potential for the higher order ABC shot (or 3-shot) is computed from the constituent 2-shots AB and BC.}
    \label{fig:higher_order}
\end{figure}

If the gaze is equally distributed among the two constituent 1-shots, the resulting 2-shot will have a high gaze potential, implying that the combined shot of the two actors has more value. Conversely, if the gaze is focused on only one of the 1-shots the 2-shot gaze potential will be lower, implying that the combined shot is less valuable. A similar hierarchy can be followed for computing the gaze potential for higher-order shots. For instance, gaze potentials of two 2-shots $G(s^{ab})$ and $G(s^{bc})$ can be used to compute gaze potential of a 3-shot $G(s^{abc})$, when the actors appear on screen in the order $a,b,c$ on moving left to right. Gaze potential computation is illustrated in Figures~\ref{fig:gaze_potential2} and~\ref{fig:higher_order}.  Fig.~\ref{fig:gaze_potential2} presents gaze potential computation for a 2-shot, while Fig.~\ref{fig:higher_order} shows computation of a 3-shot's gaze potential from two 2-shots.





\subsubsection{Editing Cost}
While gaze cues enable identification of the most important shots from the viewer's perspective, a shot selection methodology that is purely guided by gaze may not be optimal. This is owing to two reasons: (1) Even high-end eye-tracking hardware is prone to noisy measurements, especially over time; (2) If users constantly keep shifting their focus between two actors, frequently cutting between these two shots may negatively impact viewing experience of the edited video.
\begin{figure*}[t!]
    \centering
   \includegraphics[width=0.85\textwidth]{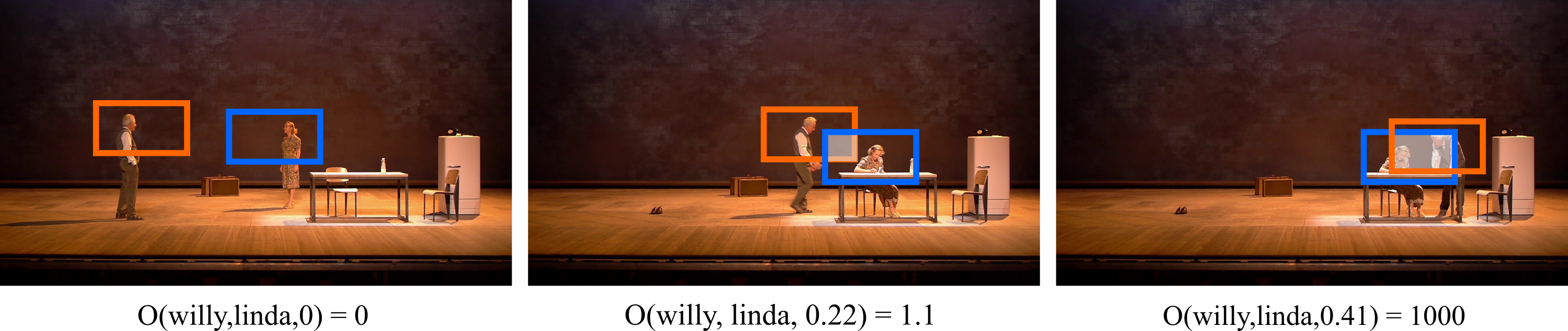} 
    \caption{ Shot-overlap cost computation: Figure shows three different master shots, and overlap cost is evaluated over switching between the two 1-shots in each case. The overlap cost $(O)$ increases from 0 to 1.1 to 1000 as IoU for the 1-shots varies from 0 to 0.22 to 0.41. In essence, jump cuts are precluded when there is significant overlap between actor shots.
    }
    \label{fig:gaze_overlap}
\end{figure*}

Therefore, an important objective of video editing is to ensure smooth shot transitions that sustain a continuous and clear narration. This is often achieved by relying on a well established set of rules such as avoiding jump cuts, maintaining left to right ordering of on-screen characters \etc. Another important element of film editing style is the cutting \textit{rhythm}, which denotes the average time duration between cuts. Cutting rhythm can be manipulated to control the style and tone of the scene. We model these important elements of the video editing process as penalty terms in the objective function denoted by Eq.~\ref{objective_function}. We introduce three types of penalty terms, namely (a) shot transition cost $(T)$, (b) shot overlap cost $(O)$ and (c) cutting rhythm cost $(R)$. The total cost for transitioning from shot $r_{t-1}$ to shot $r_{t}$ ($ r_{t-1},r_{t} \in S $) is the cumulative sum of all these costs, \ie,
\vspace{-.1cm}
\begin{equation}
    E_e(r_{t-1},r_{t}) = T(r_{t-1},r_{t}) + O(r_{t-1},r_{t},\gamma) + R(r_t,r_{t-1},\tau) 
\end{equation}

\subsubsection*{Shot transition cost}


Frequent shot transitions may not allow the viewer enough time to comprehend the scene, and rapid cuts can disrupt the viewing experience. To avoid frequent transitions from one shot to another, we introduce the transition cost. Given two shots $s_t^p$ and $s_{t+1}^q$ across consecutive time steps, the transition cost is defined as 

\vspace{-.2cm}
\begin{equation}
T(s_t^p,s_{t+1}^q) = \left\{
                        \begin{array}{lr}
                            0 & p=q\\
                             \lambda & p\neq q 
                        \end{array}
                     \right.
\end{equation}
where $\lambda$ is the transition cost parameter. The above term penalizes cuts, and motivates shot sustenance over time.

\subsubsection*{Overlap cost}

While cutting from one shot to another, the overlap between the two framings should be sufficiently low, otherwise it results in a cut that gives the effect of jumping in time, and is hence termed a \textit{jump cut}. To prevent jump cuts, we introduce a overlap cost.  
\vspace{-.1cm}
\begin{equation}
O(s_t^p,s_{t+1}^q,\gamma) = \left\{
                        \begin{array}{lr}
                            0 & \gamma \leq \alpha\\
                            \frac{\mu \gamma}{\alpha} & \alpha \le \gamma \leq \beta \\
                            \nu &  \gamma \geq \beta\\
                        \end{array}
                     \right.
\end{equation}
Here, $\gamma$ is the overlap ratio defined as the intersection over union (IoU) of two distinct shots $s_t^p,s_{t+1}^q$. The overlap cost is a piece-wise function where no cost is incurred when IoU ($\gamma$) is below threshold $\alpha$, linear cost is incurred with IoU between thresholds $\alpha$ and $\beta$, and high penalty  $\nu$ incurs when the IoU is greater than $\beta$. Figure~\ref{fig:gaze_overlap} illustrates how overlap cost varies in different scenarios.  

\subsubsection*{Rhythm cost}
Frequency of cuts plays a key role in video editing. Shot length greatly effects how a scene is perceived by the audience. Slower rhythm or longer shots invoke a sense of stillness, which is often adopted in romantic scenes where emotions have to be portrayed. Faster rhythm or shorter shots are used when high energy or chaos has to be shown; this is commonly employed in action sequences. To control the cut rhythm, we introduce a cost based on shot duration. We define rhythm cost as:

\vspace{-0.2cm}
\begin{equation}
R(s_t^p,s^q_{t-1},\tau) = \left\{
                        \begin{array}{lr}
                            \gamma_1\left( 1 -  \frac{1}{1+\exp{(l-\tau)}}\right) & 
                            p \neq q \\
                            
                            \gamma_2\left( 1 -  \frac{1}{1+\exp{(\tau-m})}\right) &  p = q\\
                        \end{array}
                     \right.
\end{equation}

\noindent where transition has been made from shot $s^p_{t-1}$ to $s^q_{t}$, upon $s^p_{t-1}$ prevailing for $\tau$ seconds. $l$ and $m$ are rhythm parameters and $\gamma_1$ and $\gamma_2$ are scaling constants. The rhythm cost is a case-wise penalty. When transition is made from a shot to itself $(p=q)$, we have a monotonically increasing penalty which becomes significant at $\tau = m$ seconds and accumulates over time. This motivates the introduction of a new cut. Once a cut occurs $(p \neq q)$, a monotonically decreasing function is triggered and adds a high penalty if another cut is introduced before $\tau = l$ seconds. The two terms together control the cutting rhythm.




\subsection{Optimizing Edits}
We solve Equation \ref{objective_function} using dynamic programming. The algorithm outputs a sequence of shots {$r_t$} for each frame $t$ selected from the set of shots generated over time $\{S_t\}$. We build a cost matrix $C(r_t,t)$ (where $r_t \in {s_t^i}$, $t = [1 .. T]$) where each cell is computed with recurrence relation resulting from Eq.~\ref{objective_function} :
\vspace{-.1cm}

\begin{equation}
\small
    C(r_t,t) = \left \{ \begin{array}{lr}
                    -ln(G(r_t)) &  t=1\\ 
                \min_{k}[C(r_{k},t-1) -ln(G(r_t)) + E_e(r_k,r_t)] & otherwise
    \end{array}
    \right \}
\end{equation}

\noindent The matrix is built along the time dimension. For each cell in the matrix, we compute and store the minimum cost to reach it. Once the matrix is built, we then perform backtracking to retrieve the sequence of optimal shots. We present the original wide angle recording during the first four seconds of the edited video as an \textit{establishing shot} and only optimize over the remaining video frames.

\subsection{Personalizing Edits}
Parameters of GAZED are either inspired from literature or empirically set. For example, transitions with less than 20\% ($\alpha = 0.2$) overlap are fluid and more than 40\% ($\beta = 0.4$) overlap appear abrupt. The rhythm parameter $m$ is set to 7 seconds based on the average shot length used in movies over two decades~\cite{cutting2015shot}. $\gamma1 = 100$ and $\nu = 1000$ are kept relatively high, to avoid extremely fast cuts and to avoid jump cuts respectively. Most parameters are generic and fixed; however, some parameters like $m$ and $l$ can be varied for personalization (faster pace or allowing shorter shots). Since our algorithm is computationally efficient, it enables interactive content exploration. In the  supplementary material, we illustrate example results of the same video edited at different pace/rhythm. 


\section{Experiments}
To examine if GAZED video editing, which incorporates both gaze information and cinematic principles results in a vivid, engaging and aesthetic rendering of a stage performance recorded with a static and distant camera, we performed a user study detailed below.

\subsection{Materials and Methods}\label{DC}
\blue{We selected a total of 12 stage performance videos for evaluation, five of them recorded at 4K resolution (3840 $\times$ 2160 pixels) and another seven in Full HD (1920 $\times$ 1080 pixels). These videos are wide-angle recordings, where  a static camera captures the entire scene-of-interest, and the videos are devoid of any pan/cut/zoom operations. The 12 videos depict a rich variety of events such as \textit{music concerts}, \textit{dance performances} and \textit{dialogue scenes} from theatrical acts. These videos were carefully selected so as to have diversity in the pace of the scenes (\eg, slow and fast dance movements, long monologues and rapid conversations), periodically necessitating pans and cuts to captivate viewer attention and present them with a detailed view of the center of action. Cumulatively, the 12 videos lasted 12 minutes and 4 seconds, with individual videos ranging from 45--80 seconds.}


\blue{The GAZED system envisages an editor/director reviewing a stage recording, and implicitly selecting the shots of interest over time via eye-gaze in lieu of manual selection. Expensive eye-trackers that accurately record eye-gaze at high speed ($\geq$ 1000 samples/second) exist today. However, we recorded gaze using a highly affordable ($<\EUR{200}$) eye-tracker in this work to demonstrate that GAZED can generate professional edits even with cheap eye tracking hardware.}
 
\blue{To compensate for tracking inaccuracies and sampling limitations of low-end eye trackers, five student users with normal or corrected vision were recruited for recording gaze. All five users were naive to the purpose of the study, and had not earlier watched any of 12 videos used in our experiments. Participants viewed the 12 videos re-sized to $1920\times1080$ pixels size on a 15.6 inch PC monitor. Viewers sat about 60 cm away from the screen, and ergonomic settings were adjusted prior to the experiment.} 

The \textit{Tobii EyeX} eye-tracker having a 60 Hz sampling rate and gaming-level eye tracking precision was used to record gaze data. The tracker was calibrated via the 9-point method before recording began. MATLAB PsychToolbox~\cite{brainard07} was used to develop the video presentation and gaze recording protocol. The 12 videos were presented to users in a fixed order for gaze recording.


\subsection{Video Editing Baselines}

\blue{As outlined previously, \textit{video retargeting} and \textit{video editing} are two different problems. As illustrated in Fig.~\ref{retargeting-vs-editing}, \textit{video retargeting} only involves estimation of a 
center of the cropping window location (single x-coordinate) within the original video at each time. In contrast \textit{video editing} requires generation of multiple virtual shots with desired compositions (parameterized by $x,y$ and zoom value at each frame) followed by shot selection. Therefore, a comparison between the GAZED editing algorithm and state-of-the-art gaze-based video retargeting methods~\cite{jain2015gaze,rachavarapu2018watch} would neither be meaningful nor fair.} We instead compare GAZED against four competent video editing baselines, namely, \textit{random}, \textit{wide}, \textit{greedy gaze} and \textit{speaker-based}, which are described below. \blue{Edits obtained via these strategies for exemplar recordings are presented in the supplementary video}. For fair comparison, the \textit{minimum shot duration} parameter $l$ is set to 1.5 seconds, and the initial scene is established via presentation of the \textit{master shots} (original video frames) for the first four seconds for all methods (except wide baseline).


\subsubsection{Random}
The Random (Ran) baseline is the weakest or the most context-insensitive video editing method. In this approach, one shot among the multiple rushes is arbitrarily selected for rendering every $l$ seconds independent of the scene information. 

\subsubsection{Wide}
The Wide baseline is motivated by the idea of video retargeting, and is in principle equivalent to the letterboxing method described in~\cite{jain2015gaze}. In the Wide shot selection strategy, the widest possible shot covering all stage performers is always preferred. This wide shot represents a {zoomed-in} version of the master shot, and denotes the smallest bounding box that covers all performers. As an illustrative example, the `ABC' shot in Figure~\ref{fig:shot_generation} denotes the Wide baseline. 

\begin{figure*}[htbp]
    \centering
   \includegraphics[width=0.248\linewidth]{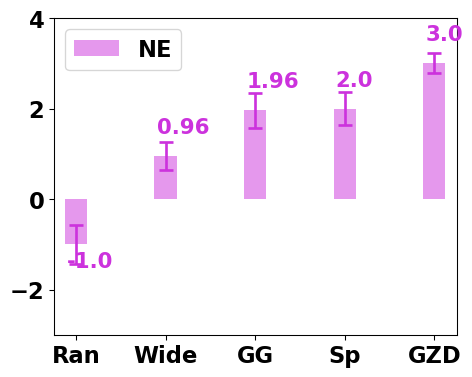}\hspace{0.01cm}\includegraphics[width=0.248\linewidth]{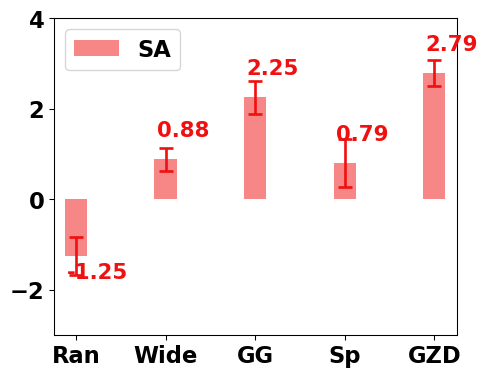}\hspace{0.01cm}\includegraphics[width=0.248\linewidth]{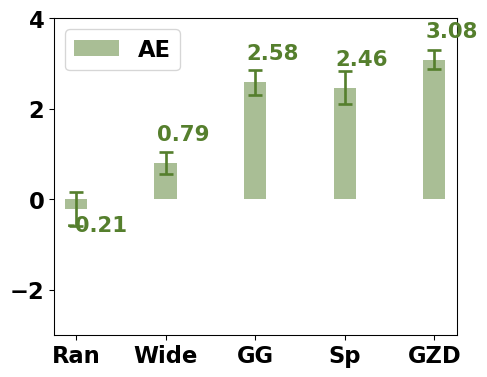}\hspace{0.01cm}\includegraphics[width=0.248\linewidth]{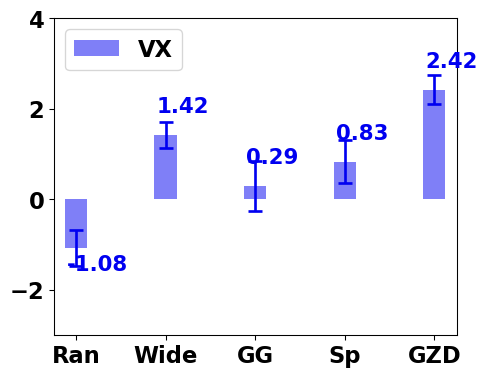}
    \caption{Bar plots denoting mean user ratings for the different evaluation attributes and editing methodologies. Error bars denote standard error of mean. Best viewed in color and under zoom. }
    \label{fig:US_plots}
\end{figure*}

\blue{\subsubsection{Greedy gaze}
The greedy gaze (GG) editing algorithm greedily selects the shot with \textit{maximum} gaze potential to render at every time instant. In terms of shot selection, the main difference between the GG baseline and GAZED is as follows: while GAZED performs a \textit{global} optimization by minimizing Equation~\ref{objective_function} to derive the optimal sequence of shots $\{r_t\}$ to present over time, the GG baseline directly selects the presentation shot $r_t$ based on the \textit{local gaze potential optimum} at time $t$. In addition, since this editing strategy is solely guided by gaze information without adhering to cinematic editing guidelines, frequent switching of shots may occur which would negatively impact comprehension of the scene content and consequently, the viewing experience. To preclude the occurrence of transient shots, we impose a minimum shot duration of 1.5 seconds as specified by the the $l$ parameter. }

\subsubsection{Speaker-based}
Speaker cues have proved useful for editing dialog-driven scenes~\cite{ranjan2008improving, leake2017computational}. Ranjal \etal~\cite{ranjan2008improving} and Leake~\etal~\cite{leake2017computational} favor selection of shots where the speaker is clearly visible. Our speaker-based (Sp) editing baseline similarly selects the 1-shot that best captures the speaker from among the rushes. For the purpose of this work, speaker information in each video was annotated manually. When more than one person speaks simultaneously, their combined shot is selected. The algorithm persists with its current selection till a change of speaker occurs. A minimum shot duration (specified by $l$) is enforced to avoid rapid shot transitions. If silence of more than 10 seconds is observed, the \textit{wide} shot is selected for the next time instant. 

\subsection{User Study}\label{US}

\begin{sloppypar}
To evaluate the video editing qualities of GAZED (GZD) against the above baselines, we conducted a psychophysical study with 12 users (different from those used for compiling gaze data) and the 12 video recordings described earlier. Edited versions of the 12 videos were generated via the four baseline plus GAZED strategies (for fair comparison, all parameters of GAZED were kept same across all videos). Upon viewing the original video, each user viewed the five edited video versions in random order on the same PC screen used for gaze recording. We designed the study such that each user viewed the original and edited versions of two stage recordings, so that six users cumulatively viewed all the 12 recordings. This resulted in a $12~(\text{video types}) \times \text{2~(user ratings/video)} \times 5~\text{(editing strategies)}$ factor design. 
\end{sloppypar}

\blue{Users were naive to the strategy employed for generating each edited version that they watched. On viewing each edited version of the original, users had to \textit{compare} the edited version against the original, and provide a Likert rating on a [-5,5] scale for each of the attributes described below. These attributes were adopted from the retargeting study described in~\cite{rachavarapu2018watch}-- we note here that while video editing and retargeting are technically different, they are nevertheless designed to direct the viewer's attention onto \textit{focal scene events} given rendering constraints. Therefore, psychophysical questions posed for evaluating video retargeting methods are also relevant for video editing. Attributes of interest included:}
 
\begin{itemize}
 \item[(1)] \textbf{Narrational Effectiveness (NE):} \textit{How effectively did the edited video convey the original narrative?}
 \item[(2)] \textbf{Scene actions (SA):} \textit{How well did the edited video capture actor movements and actions?}
 \item[(3)] \textbf{Actor Emotions (AE):} \textit{How well did the edited video capture actor emotions?}
 \item[(4)] \textbf{Viewing experience (VX):} \textit{How would you rate the edited video for aesthetic quality?}
\end{itemize}

Users were educated regarding these attributes, and about cinematic video editing conventions prior to the study. Users had to rate for questions (1)--(4), \textit{relative} to a reference score of `0' for the original video. A \textit{positive} score would therefore imply that the edited version was \textit{better} than the original for the attribute in question, while a \textit{negative} score conveyed that the edited version was \textit{worse} than the original with respect to the criterion. User responses were collated, and mean scores were computed for each criterion and editing strategy over all videos (see Figure~\ref{fig:US_plots}). Statistics and inferences from the user study are presented below.

\subsubsection{Results and Discussion}
\blue{A two-way balanced analysis of variance (ANOVA) on the compiled NE, SA, AE and VX user scores revealed the main effect of \textit{editing strategy} on user opinions ($p < 0.000001$ for all four attributes), but no effect of \textit{video type}. An interaction effect was also noted for AE scores ($p < 0.005$). We hypothesized that incorporating gaze cues plus cinematic rules for shot selection would generate an engaging, vivid and aesthetic edited output; Figure~\ref{fig:US_plots} validates our hypothesis as the GAZED editing strategy elicited the highest user scores for all the four considered attributes.}

\blue{Examining individual attributes, \textit{post-hoc} independent $t$-tests on NE scores revealed a significant difference between GAZED vs. Speech ($p < 0.05$), GAZED vs. GG ($p < 0.05$), GAZED vs. Wide ($p < 0.000005$) and GAZED vs. Random ($p < 0.000001$). Differences were also noted between Random and the remaining baselines ($p < 0.000005$ for Speech, $p < 0.00001$ for GG and $p < 0.001$ for Wide), and between Speech vs. Wide ($p < 0.05$). These results cumulatively convey that carefully compositing shots that provide a close-up view of the focal actor(s) and action(s) is critical for effective narration. The GG, Sp and GAZED strategies that are designed to focus on actors and actions deemed important via speech or gaze-based cues achieve better scores than the Ran and Wide baselines which can only achieve incorrect/inefficient framing of the scene characters. The Random baseline which selects shots independent of scene content performs worst, while a vivid presentation of gaze-encoded salient scene events via GAZED is perceived as the best narrative.} 

\blue{With respect to conveying scene actions, GAZED performs significantly better than Speech-based ($p < 0.005$), Wide ($p < 0.00001$) and Random ($p < 0.000001$). Also greedy gaze achieves significantly higher scores than both Speech ($p < 0.05$) and Wide ($p < 0.005$). The Wide and Sp baselines perform similarly for this criterion, while Ran performs the worst as expected. These results point to the fact that speaker cues may not be as effective for conveying focal events in stage performances; this is quite plausible as a particular actor may assume the role of a narrator, while another performs the actions of interest. Alternatively, one performer may verbally introduce co-performers to the audience, in which case the Sp baseline would still focus on the introducer instead of the introducee. In such cases therefore, eye gaze is more accurate at capturing events and actors of interest than speech. The Wide baseline, which captures the entire scene context at all times and thereby can only present low-resolution views of performers to viewers, performs similar to Sp. The GG strategy which vividly captures events of maximal interest at each time instant is nevertheless effective at conveying scene actions and performs second best. }

\blue{Considering how the different editing strategies convey actor emotions, we observe that the GG, Sp and GAZED methods perform fairly similarly. These three approaches perform significantly better than the Wide and Ran baselines ($p<0.0005$ for all pairwise comparisons). Wide still performs better than Random editing ($p<0.05$). These observations can be explained as follows-- wherever appropriate, the GG, Sp and GAZED editing methods capture the speaker or focal actor in the scene as a close-up shot, which enables an effective rendering of facial emotions to the viewer. The Wide baseline captures all scene actors at each video frame, and can therefore only present facial movements at (relatively) low-resolution to  viewers.} 

\blue{Cinematic editing principles constitute a key component of the GAZED shot selection process-- incorporating them in the objective function defined by Equation~\ref{objective_function} is critical for producing a seamless, smooth, visually engaging and aesthetic edited output. Specifically, including the shot transition and overlap costs would minimize  the number of \textit{cuts} in the edited video, and frequent cuts can distort viewer comprehension of the scene context (\eg, knowledge of scene actor locations), and thereby negatively impact viewing experience. Note from the editing baseline descriptions that none of them incorporate these editing principles for shot selection, except for enforcement of a minimum shot duration of 1.5 seconds via the $l$ parameter to preclude transient shots.} 

\blue{We hypothesized that the incorporation of cinematic editing rules in the shot selection framework, coupled with the capturing of focal scene events would maximize viewing experience of the edited video. Consistent with our expectation, GAZED scores highest among the five methods for viewing experience, and performs significantly better than the Sp ($p<0.01$), GG ($p<0.005$), Wide ($p<0.05$) and Ran ($p < 0.000001$) baselines. Wide performs second best scoring marginally better than GG ($p=0.0752$) and significantly better than Ran ($p<0.00001$), but only comparable to Sp. The superiority of Wide over GG and Sp can be attributed to the fact that the entire scene context is always visible to the viewer at all times in this editing strategy; Both Sp and GG are designed to cut routinely to focus on the (sometimes incorrectly) perceived action of interest in the scene, and such frequent cutting can lead to a jarring viewing experience. In contrast, Wide presents a relatively consistent framing of the scene to viewers at all times, resulting in a \textit{smooth} edited video. Finally, both Sp and GG elicit a better viewing experience than Ran ($p < 0.05$ in either case).}

\section{Summary and Conclusion}

This work presents the GAZED framework for automatically editing stage performance videos captured using a single, static, wide-angle and high-resolution camera employing user gaze cues. As human eyes are known to be sensitive to focal scene events, GAZED translates user gaze data into \textit{gaze potentials}, which quantify the \textit{importance} of the multiple rushes generated from the master shot via the shot generation process; the gaze potential serves a natural measure for guiding \textit{shot selection}, and the shot selection process directly impacts the quality of the edited video. While GAZED shot selections are primarily driven by user attention, cinematic editing principles such as \textit{avoiding jump cuts}, \textit{precluding transient shots} and \textit{controlling shot-change rhythm} are also modeled within an energy minimization function guiding the shot selection process. These cinematic rules facilitate the generation of a smooth, vivid, visually engaging and aesthetic output as confirmed by user opinions compiled from a psychophysical study.

Our user study compares the GAZED framework against four editing baselines, namely, \textit{Random}, \textit{Wide}, \textit{Greedy Gaze} and \textit{Speech-based}. While the Random baseline is employed to demonstrate that editing strategies should be guided by scene content, others are inspired by prior literature. The Wide baseline is motivated by letterboxing for video targeting; the speech baseline is included to mimic~\cite{ranjan2008improving} and~\cite{leake2017computational} for editing stage recordings. Greedy gaze-based editing is used to showcase how modeling cinematic editing rules can generate a smooth and aesthetic edited video. 

The user study reveals salient aspects of the different editing strategies. GAZED performs best, while Random performs worst with respect to four essential attributes of the edited video. Speech-based editing scores highly with respect to conveying actor emotions, but is perceived as ineffective at capturing focal scene actions; this observation reveals that speech cannot effectively reveal the salient actors/actions in stage performances. GG editing conveys both scene actions and action emotions convincingly, validating the utility of gaze cues for encoding focal scene events. However, it does not incorporate cinematic rules, which results in a poor viewing experience. Conversely, the Wide baseline scores low with respect to the NE, SA and AE attributes as it does not focus on individual actors. However, it consistently frames the entire scene resulting relatively smooth edited output, which elicits a moderate viewing experience.

The computational efficiency of the algorithm allows for efficient exploration and creating personalized edits. GAZED edits a minute long video (24 fps) with three performers in 1.5 seconds. The computation only grows linearly with video length and editing a 30 minute video with three performers take
s about 49 seconds. With four actors GAZED takes about 5 seconds for editing a minute long video and 162 seconds for a 30 minute video. These computations are done a PC with 7th generation Intel 2.7 GHz i5 processor and 8GB RAM. The exhaustive list of possible shots will grow exponentially with increasing number of actors, which will significantly increase the computation. However, not all these shots may be useful to convey the scene and the user (editor) can pick the relevant set of shots to select from and generate the final editing.

Limitations of the current GAZED implementation include (1) the inability to \textit{pan} between shots as the shot selection process only induces \textit{cuts} in the edited video; (2) being able to achieve only a single value of \textit{zoom} corresponding to a medium/medium close-up compositing for the 1-shots, and (3) noise induced by the eye-tracking hardware which can be largely alleviated through the user of high-end trackers, and recording gaze in bursts to enable {drift correction}. Achieving the capability to pan between shots, and  gradually zoom on faces would be the focus of future work.

Nevertheless, the utility of GAZED can be appreciated with an understanding of the complexity of the video editing process, which is extremely tedious and effort intensive. In this regard, speech-based editing~\cite{leake2017computational} which (a) is specific to dialog scenes, (b) requires the film script as input and (c) takes about 110-217 minutes for scene pre-processing, still represents a utility tool. Similarly, the GAZED system can be used by editors to inexpensively and quickly obtain the first-cut edit of performance videos and also explore different editing styles (\eg, by varying the rhythm parameter), so that they can invest more time and effort in devising refinements and creative edits where necessary.

\section{Acknowledgement}
This work was supported in part by Early Career Research Award, ECR/2017/001242, from Science and Engineering Research Board (SERB), Department of Science \& Technology, Government of India. Special thanks to Remi Ronfard, Claudia Stavisky, Auxane Dutronc and the cast and crew of ‘Death of a salesman’. 

\balance{}

\bibliographystyle{SIGCHI-Reference-Format}
\bibliography{sample}

\end{document}